\pdfoutput=1
\documentclass[11pt]{article}

\usepackage[]{ACL2023}

\usepackage{times}
\usepackage{latexsym}
\usepackage{booktabs}
\usepackage{arydshln}

\usepackage[T1]{fontenc}
\usepackage[utf8]{inputenc}

\usepackage{microtype}

\usepackage{inconsolata}
\usepackage{graphicx}
\usepackage{multirow}
\usepackage{subfig}

\title{Dialogue Quality and Emotion Annotations for Customer Support Conversations}

\author{John Mendonça\textsuperscript{1,2,*\textdagger}, Patrícia Pereira\textsuperscript{1,2,*}\\ \textbf{Miguel Menezes\textsuperscript{1,3,4}, Vera Cabarrão\textsuperscript{4}, Ana C. Farinha\textsuperscript{4}, Helena Moniz\textsuperscript{1,3,4}}\\ \textbf{João Paulo Carvalho\textsuperscript{1,2}, Alon Lavie\textsuperscript{5,6}\and Isabel Trancoso\textsuperscript{1,2}} \\
  \textsuperscript{1} INESC-ID, Lisbon \\
  \textsuperscript{2} Instituto Superior Técnico, University of Lisbon \\
  \textsuperscript{3} Faculdade de Letras, University of Lisbon \\
  \textsuperscript{4} Unbabel, Lisbon \\
  \textsuperscript{5} Carnegie Mellon University, Pittsburgh \\
  \textsuperscript{6} Phrase, Pittsburgh \\
  \texttt{\{john.mendonca, patricia.pereira\}@inesc-id.pt} \\}

\begin{document}
\maketitle
\begin{abstract}
Task-oriented conversational datasets often lack topic variability and linguistic diversity. However, with the advent of Large Language Models (LLMs) pretrained on extensive, multilingual and diverse text data, these limitations seem overcome. Nevertheless, their generalisability to different languages and domains in dialogue applications remains uncertain without benchmarking datasets. This paper presents a holistic annotation approach for emotion and conversational quality in the context of bilingual customer support conversations. By performing annotations that take into consideration the complete instances that compose a conversation, one can form a broader perspective of the dialogue as a whole. Furthermore, it provides a unique and valuable resource for the development of text classification models. To this end, we present benchmarks for Emotion Recognition and Dialogue Quality Estimation and show that further research is needed to leverage these models in a production setting.

\end{abstract}

\begingroup\def\thefootnote{*}\footnotetext{Joint first authors.}\endgroup
\begingroup\def\thefootnote{\textdagger}\footnotetext{Work partially conducted as a visiting scholar at CMU.}\endgroup

\section{Introduction}
\label{sec:intro}

\begin{table}[t!]
\centering
\small
\begin{tabular}{l}
\toprule
\cellcolor[HTML]{FDEDEC}\textbf{Agent:} Delivery usually takes place within 1-7 working\\ \cellcolor[HTML]{FDEDEC}days after dispatch but this can vary depending on the\\ \cellcolor[HTML]{FDEDEC}couriers availability in your area.\\
\textbf{Cor:} 2	\hspace{0.3cm}\textbf{Tem:} 1	\hspace{0.3cm}\textbf{Eng:} 1 \hspace{0.3cm} \textbf{Emo:} \textit{Neutral} \\ 
\textbf{Und:} 1	\hspace{0.3cm}\textbf{Sen:} 1	\hspace{0.4cm}\textbf{IQ:} 5 \hspace{0.6cm}\textbf{Pol:} 1 \\\midrule\midrule

\hspace{1.9cm}\textbf{Customer:}\cellcolor[HTML]{ECF4FF} Sorry, that doesn't satisfy me.\\
\textbf{Cor:} 2	\hspace{0.3cm}\textbf{Tem:} 0	\hspace{0.3cm}\textbf{Eng:} 1 \hspace{0.3cm}\textbf{Emo:} \textit{Anger}\\ \midrule
\hspace{0.5cm}\cellcolor[HTML]{ECF4FF}I'm already waiting for my sofa for almost 8 weeks!\\
\textbf{Cor:} 2	\hspace{0.3cm}\textbf{Tem:} 0	\hspace{0.3cm}\textbf{Eng:} 1 \hspace{0.3cm}\textbf{Emo:} \textit{Anxiety}\\ 
\textbf{Und:} 1	\hspace{0.3cm}\textbf{Sen:} 1	\hspace{0.4cm}\textbf{IQ:} 1 \hspace{0.5cm}\textbf{Pol:} 0 \\\midrule\midrule

\textbf{A:} \cellcolor[HTML]{FDEDEC}Regrettably as we do not have control over the speeds\\ \cellcolor[HTML]{FDEDEC}of the shipping processes we are unable to expedite\\ \cellcolor[HTML]{FDEDEC}orders, the item is still on a boat but we are doing our\\ \cellcolor[HTML]{FDEDEC}best to get it to you as soon as possible.\\
\textbf{Cor:} 2	\hspace{0.3cm}\textbf{Tem:} 0	\hspace{0.3cm}\textbf{Eng:} 0 \hspace{0.3cm}\textbf{Emo:} \textit{Disappointment}\\ \midrule

\cellcolor[HTML]{FDEDEC}Any delays the item may encounter on its way to our\\\cellcolor[HTML]{FDEDEC}distribution center are out of our hands and cannot be \\\cellcolor[HTML]{FDEDEC}predicted.\\
\textbf{Cor:} 2	\hspace{0.3cm}\textbf{Tem:} 0	\hspace{0.3cm}\textbf{Eng:} 0 \hspace{0.3cm}\textbf{Emo:} \textit{Neutral}\\ 
\textbf{Und:} 2	\hspace{0.3cm}\textbf{Sen:} 1	\hspace{0.4cm}\textbf{IQ:} 5 \hspace{0.5cm}\textbf{Pol:} 1\\\midrule\midrule

\hspace{1.5cm}\textbf{C:} \cellcolor[HTML]{ECF4FF}And now? Should I sit or lie on the floor? \\
\textbf{Cor:} 2	\hspace{0.3cm}\textbf{Tem:} 0	\hspace{0.3cm}\textbf{Eng:} 1 \hspace{0.3cm}\textbf{Emo:} \textit{Frustration}\\ \midrule
\hspace{5.1cm}\cellcolor[HTML]{ECF4FF}This is not okay\\
\textbf{Cor:} 2	\hspace{0.3cm}\textbf{Tem:} 0	\hspace{0.3cm}\textbf{Eng:} 1 \hspace{0.3cm}\textbf{Emo:} \textit{Frustration}\\ 
\textbf{Und:} 1	\hspace{0.3cm}\textbf{Sen:} 1	\hspace{0.4cm}\textbf{IQ:} 1 \hspace{0.5cm}\textbf{Pol:} 0 \\\midrule\midrule

\textbf{A:}\cellcolor[HTML]{FDEDEC} I understand this is frustrating and disappointing.\\
\textbf{Cor:} 2	\hspace{0.3cm}\textbf{Tem:} 1	\hspace{0.3cm}\textbf{Eng:} 1 \hspace{0.3cm}\textbf{Emo:} \textit{Neutral}\\
\textbf{Und:} 2	\hspace{0.3cm}\textbf{Sen:} 1	\hspace{0.4cm}\textbf{IQ:} 5 \hspace{0.5cm}\textbf{Pol:} 1\\\bottomrule

\end{tabular}
\caption{Adapted example of a portion of a dialogue from the MAIA DE-1 subset, from the point of view of the Agent (which receives and sends messages in English). The customer interacts with the agent in their corresponding language (in this case German). This is achieved by employing Machine Translation on both ends (DE $\rightarrow$ EN and EN $\rightarrow$ DE).}
\label{tab:dialogue}
\end{table}

% Miguel
%Artificial Intelligence (AI) is very much an omnipresent technology. This powerhouse state-of-the-art technology is, nevertheless, very much dependent on data, hence the epithet data-driven technologies. However, to fully ripe AI´s full potential, the data used to propel this type of technologies must be labelled according to needs and goals. This operation, often referred as data annotation, provides the ground-truth for Machine Learning (ML), impacting and influencing algorithm performance.

Artificial Intelligence (AI) has evolved to become a ubiquitous technology in our lives. Yet, its performance is limited by the amount of data it is trained on. Therefore, and in order to maximise the rewards of such technology, substantial research and engineering effort has been devoted to collecting and annotating data according to needs and goals.

One of the main limitations of most task-oriented conversational datasets is their lack of variability. The majority of these datasets are collected in controlled environments where annotators are encouraged to follow specific guidelines, and are limited to a restrictive set of topics, and outcomes \cite{el-asri-etal-2017-frames,budzianowski-etal-2018-multiwoz,rastogi2020towards}. This leads to highly structured dialogues that do not accurately reflect genuine conversations. In contrast, customer support conversations provide a broader range of topics and contexts, and are more linguistically diverse \cite{lowe-etal-2015-ubuntu}. Furthermore, most datasets are monolingual, resulting in a lack of representation of diverse linguistic and cultural features such as tone and idiomatic expressions \cite{goncalo-oliveira-etal-2022-brief}.

One approach to equip NLP models with multilingual and diverse domain knowledge capabilities is to leverage LLMs pretrained on extensive amounts of publicly available data \cite{conneau-etal-2020-unsupervised, xue-etal-2021-mt5,openai2023gpt4}. However, lacking benchmarking dialogue datasets, it is not clear these models, applied to dialogue, are able to fully generalise to other languages and/or domains, even if other dimensions of variability remain unchanged.

\textbf{This paper builds upon the original MAIA dataset} release by adding extensive annotations of emotion and dialogue quality at different granularity levels, thus allowing a holistic approach at understanding the dynamics of conversations in the context of customer support. The MAIA dataset is a collection of genuine bilingual customer support conversations initially released as a challenge dataset for the WMT Chat shared-task \cite{farinha-etal-2022-findings}. In these conversations, which are powered by Machine Translation, the agent communicates with the customer exclusively in English, whereas the customer interacts with the agent exclusively in their native language. Our annotations cover 612 dialogues accounting for around 25k sentences, covering diverse topics, ranging from account registration issues, payment and delivery clarifications and aftersale services. Languages includes German (DE), Brazilian-Portuguese (PT\_BR) and European Portuguese (PT\_PT).

We argue that the MAIA dataset and the accompanying annotations have unique value in the field of customer support and conversational agents. The comprehensive annotations conducted enable the analysis of the relations between several dialogue sub-qualities and emotion. Furthermore, they can be used as a training and benchmark dataset for text classification in these distinctive settings. For instance, one could leverage this dataset for the construction of dialogue systems that support customer-agent interaction processes.
Classification models trained on this data could assist customer service agents (human or machine) by measuring customer emotions and dialogue qualities in real-time and provide the agent with feedback on the fluidity and success of the dialog. 

To kick-start this research, this paper provides benchmarks for Emotion Recognition and Dialogue Quality Estimation. Results show that existing models are not strong enough to perform on par with other benchmarks, indicating significant future work research will be required to reduce this performance gap.

In summary, the primary contributions of this work are as follow:

\begin{itemize}
    \item We conduct extensive emotion and dialogue quality annotations for the MAIA dataset. More specifically, we annotate the dataset at different levels of granularity, ranging from the \textbf{sentence level}, where we perform an 8-class emotion and local conversational quality annotations; \textbf{turn-level} conversational quality annotations including IQ (Interaction Quality); up to the \textbf{dialogue level}, where annotations for task success are provided;
    \item We analyse these annotations and show how emotions and different aspects of conversational quality are related in the context of customer support;
    \item We benchmark known approaches for Emotion Recognition in Conversations and Dialogue Evaluation on this dataset;
    \item The corpus and accompanying benchmarks are publicly available under the Creative Commons public license Attribution-NoDerivatives 4.0 International (CC BY-ND 4.0) and can be freely used for research purposes only. \footnote{\url{github.com/johndmendonca/MAIA-DQE}}
\end{itemize}

The paper is structured as follows: Section \ref{sec:related_work} provides a brief literature review on task-oriented dialogues and their annotations. In Section \ref{sec:processing}, the MAIA dataset construction pipeline is presented, including the anonymization and annotation steps. The dataset is formally presented in Section \ref{sec:description}, delving into the uniqueness of the dataset and its contributions to research. Existing AI-powered approaches for customer support chat such as Emotion Recognition in Conversations and Dialogue Evaluation are benchmarked in Section \ref{sec:benchmark}.

\section{Related Work}
\label{sec:related_work}

\subsection{Task-oriented Dialogue Datasets}

Perhaps the most well known open-source customer support datasets are TweetSumm \cite{feigenblat-etal-2021-tweetsumm-dialog} and the Ubuntu Dialogue Corpus \cite{lowe-etal-2015-ubuntu}. In both datasets, the language used is exclusively English. TweetSumm contains customer support interactions between customers and companies crawled from Twitter, whereas Ubuntu extracts its dialogues from the Ubuntu chat logs. The main difference between the Ubuntu dataset and TweetSum is the fact the former is constrained by the nature of the platform itself, typically resulting in limited turn interactions where the agent inevitably steers the customer to a dedicated costumer service chat platform. The Ubuntu dataset, similarly to MAIA, does not have this limitation and consists of live multi-turn dyadic conversations. However, unlike Ubuntu, the MAIA dataset contains customer support conversations of 4 different products and companies, where the agent is a representative of the company. This contrasts with Ubuntu, where the participant offering support is typically an experienced user without any official affiliation with Ubuntu. As such, the conversational dynamics between the two datasets are quite different, with the MAIA dataset showing more diverse emotions.

Other relevant public resources of task-oriented dialogue corpora include the MultiWoz and associated datasets \cite{budzianowski-etal-2018-multiwoz}. These datasets are frequently used in the context of task-oriented dialogue, where an agent assists a customer in well defined tasks such as reservations. Unlike the MAIA dataset, the interactions are collected using English speaking crowdworkers, lacking representation of other languages. Additionally, the strict guidelines result in "sterile" and structured interactions that lack complexity known to real-world customer support interactions.

\subsection{Dialogue Annotations}

One of the most widely used dialogue benchmark datasets with emotion annotations is DailyDialog \cite{li-etal-2017-dailydialog}, built from websites used to practice English and labelled with the six Ekman’s basic emotions \cite{ekman1999basic}.
In the realm of customer support, \citet{herzig-etal-2016-classifying} collected and annotated data in terms of emotions from two North America based customer support Twitter accounts. A particularity of this work is that a different set
of emotion classes was used for the agent and customer. Furthermore, annotators were asked to indicate the intensity of each possible emotion, allowing for a multi-class setting.

With respect to quality annotations, the goal of most human annotation work is to evaluate dialogue systems or to validate proposed automated metrics. As such, two approaches are typically employed: annotators either interact with the system in a live setting and rate it, or evaluate existing responses given a context which was fed to the system. In the context of task-oriented dialogue, annotating Task Success \cite{walker-etal-1997-paradise}, User Satisfaction and/or Emotion \cite{schmitt-etal-2012-parameterized} are the norm. However, for open-domain dialogue, the focus has been mostly on annotating system responses on several notions of quality \cite{see-etal-2019-makes,mehri-eskenazi-2020-unsupervised}, since these dialogues are open in nature. To the best of our knowledge, this work is the first one to provide human judgements of customer support conversations with both task-oriented and open domain dialogue quality annotations at the turn and dialogue-level.

\section{Processing and Annotations}
\label{sec:processing}

\subsection{Collection and anonymization}

The conversations that compose this corpus are extracted from the original WMT22 Chat shared-task dataset \cite{farinha-etal-2022-findings}. It consists of dialogues obtained from companies that provide customer support and that gave written consent to use their data for research purposes
%, as long 
\footnote{In accordance with the EU General Data Protection Regulation (GDPR).}. This was achieved by using a mix of proprietary anonymization tools and human annotations was used to anonymize all PII (Personally Identifiable Information) from the data\footnote{Additional information, including the anonymization tokens, are available in the original paper.}.

%In order to comply with the General Data Protection Regulation (GDPR) ~\cite{regulation2018general} and the European AI Act ~\cite{madiega2021artificial}, we  anonymized all Personally Identifiable Information (PII) from the data. For this, we first used a proprietary anonymization tool to automatically anonymize them, and then manually verified the data. This process resulted in 12 different categories of anonymization, each represented by a specific token, which can be found in Table \ref{tab:tokens-anonymization}. 
%It's worth noting that Unbabel is also certified for ISO/IEC 27001:2013 Information Security Management Certification\footnote{\url{https://resources.unbabel.com/blog/unbabel-awarded-iso-iec-27001-2013-infor\\mation-security-management-certification}}. 

%\begin{table}[!ht]
%\centering
%\small
%\setlength{\tabcolsep}{.4em}
%\begin{tabular}{ll}\\ \toprule
%\textbf{Token} & \textbf{Description}  \\  \midrule
%\textbf{\#NAME\#} & Person’s names  \\ 
%\textbf{\#PRS\_ORG\#}  & Products, Services, \\
% & and Organizations \\
%\textbf{\#ADDRESS\#}   & Address        \\
%\textbf{\#EMAIL\#}   & E-mail address \\
%\textbf{\#IP\#}   & 	IP Address        \\
%\textbf{\#PASSWORD\#}   & 	Password        \\
%\textbf{\#PHONENUMBER\#}   & Phone number        \\
% \textbf{\#CREDITCARD\#}  & 	Credit card number  \\
% \textbf{\#URL\#}  & 	URL Address        \\
% \textbf{\#IBAN\#}  & 	IBAN Address        \\
% \textbf{\#NUMBER\#}  & 	Any number (all digits)  \\
% \textbf{\#ALPHANUMERIC\_ID\#}  & 	Any alphanumeric ID     \\\bottomrule
%\end{tabular}
%\caption{Anonymization tokens and their description.}
%\label{tab:tokens-anonymization}
%\end{table}

\subsection{Annotations}

The annotations were conducted by expert linguists in the given language. A single annotator for each language was used to fully annotate the dataset. Given its structure, we annotated the dataset along three dimensions: \textbf{Sentence level}: corresponding to a single message; 
\textbf{Turn level}: one or more sentences sent by one of the participants within a given time frame. \textbf{Dialogue level}: a succession of turns between the customer and agent denoting the full conversation. Considering dialogues are collaborative acts between speakers, we annotated data from both participants, customer and agent. This allowed us to evaluate the interaction as a whole and understand how one's action may impact the following response and how that affects the outcome of the conversation. A fully annotated dialogue is presented in Appendix \ref{sec:dial}.

\subsubsection{Sentence Level Evaluation}

The metrics used to assess each sentence are as follows:
\begin{itemize}
  \setlength{\itemsep}{1pt}
  \setlength{\parskip}{0pt}
  \setlength{\parsep}{0pt}
    \item \textbf{Correctness \{0,1,2\}}
    \item \textbf{Templated \{0,1\}}
    \item \textbf{Engagement \{0,1\}}
\end{itemize}

The \textbf{Correctness} metric was expressed resorting to three different scores measuring the sentence fluency. A score of 0 applies to a sentence indicated ungrammaticalities at several levels, both in terms of structure and in terms of orthography, originating a sentence that is difficult to understand. A score of 1 indicates that the analysed sentence contains minor mistakes but still remains fully understandable. A score of 2 was used when the sentence showed no mistakes and was fully understandable and coherent.

The \textbf{Templated} metric measured the type of sentence. For each sentence, a score of 0 was given for non-templated sentences, and a score of 1 for templated sentences. Note that by templated sentences we refer to predefined scripts used by customer support agents.

The \textbf{Engagement} metric was also expressed as one of two scores, measuring the level of engagement from both conversation parties. A score of 0 indicates a lack of engagement, whereas with a score 1 the participant was fully engaged in the conversation. 

Besides the above-mentioned metrics, we also found to be reasonable to measure real emotions that usually go hand in hand within a customer support scenario. Following the previous strategy, the assessment was provided at a sentence-level, identifying the emotions conveyed by each sentence. The set of emotions used are as follows:
\textbf{Happiness; Empathy; Neutral; Disappointment; Confusion; Frustration; Anger; and Anxiety}. We selected these emotions because upon analyzing the dataset we observed that these were the most common emotions displayed from a pool of several customer support emotions. With regards to empathy, it is a crucial emotion to analyze to measure agent performance. In terms of emotion annotation, and since a situation often triggers multiple emotions, annotators had the opportunity to select multiple emotions for a single sentence, ranking from the main emotion expressed to the others that are less evident. For example, a customer can be both disappointed and frustrated.

\begin{table*}[ht]
\centering
\small
\begin{tabular}{l|c|ccc|cccc|cc}
\multicolumn{1}{c|}{}                        & \multicolumn{4}{c|}{\textbf{Sentence}}                  & \multicolumn{4}{c|}{\textbf{Turn}}                       & \multicolumn{2}{c}{\textbf{Dialogue}} \\ \hline
\multicolumn{1}{c|}{\textbf{Agreement (\%)}} & \textbf{Emot} & \textbf{Corr} & \textbf{Temp} & \textbf{Enga} & \textbf{Unde} & \textbf{Sens} & \textbf{Poli} & \textbf{IQ}    & \textbf{DC}       & \textbf{TS}       \\ \hline
\textbf{Full}                                & 72.39       & 81.45       & 76.10       & 71.24       & 88.98       & 92.12       & 98.36       & 51.97          & 90.00             & 30.00             \\
\textbf{Partial}                             & 23.06       & 17.39       & 23.90       & 28.76       & 11.02       & 7.88        & 1.64        & 41.73 & 10.00             & 60.00             \\
\textbf{None}                                & 4.56        & 1.16        & 0.00        & 0.00        & 0.00        & 0.00        & 0.00        & 6.30           & 0.00              & 10.00             \\ \hline
\end{tabular}
\caption{Observed agreement as a percentage of the total annotations per category between 3 annotators on a subset of PT\_PT-3. Annotation types are abbreviated for brevity.}
\label{tab:iaa}
\end{table*}

\begin{table*}[ht]
\centering
\small
\begin{tabular}{lccccccc}\toprule
\textbf{Metric}         & \textbf{DE-1} & \textbf{DE-2} & \textbf{PT\_BR-2} & \textbf{PT\_PT-3} & \textbf{PT\_BR-4}  & \textbf{Total} \\ \midrule
\textbf{\# Dialogues}        & 370           & 65            & 113               & 21                & 43                & 612            \\
\textbf{\# Sentences}    & 12,169        & 3,823         & 6,673             & 815               & 1,480             & 24,960         \\
\textbf{\# Tokens}      & 359,030       & 101,001       & 166,049           & 22,656            & 41,410            & 690,146        \\
\textbf{Avg. Sen/Dial}  & 32            & 58            & 59                & 38                & 34                & 40             \\
\textbf{Avg. Token/Sen} & 29            & 26            & 24                & 27                & 28                & 27             \\ \bottomrule
\end{tabular}
\caption{Statistical information of the MAIA dataset. The number of tokens includes tokens from Source and MT.}
\label{tab:stats}
\end{table*}

\subsubsection{Turn Level Evaluation}

The annotation process was designed to measure the interaction between participants within a dialogue. Since dialogues are a multi-tier architecture structure engineered not just around sentences but also around turns, it was necessary to account for these compositional properties. 
An analysis at the turn level allowed us to understand the overall mood and attitude of the turn-taker w.r.t what was previously stated by the other dialogue participant, at any given stage of the conversation. 
As a metric deeply dependent of the previously sentences, it is important to note that the initial turns were considered as non-evaluatable, since their function within the dialogue is to set the tone and the context that allow the newly started conversation to flow. 
The set of categories used for the turn taking evaluation were as follow:
\begin{itemize}
  \setlength{\itemsep}{1pt}
  \setlength{\parskip}{0pt}
  \setlength{\parsep}{0pt}
    \item \textbf{Understanding \{0,1\}}
    \item \textbf{Sensibleness \{0,1\}}
    \item \textbf{Politeness \{0,1\}}
    \item \textbf{Interaction Quality [1,5]}
\end{itemize}

The category \textbf{Understanding} measured how well the participant was able to understand the message from the other dialogue participant, with a score of 0 meaning the understandability was somehow compromised, and the score 1 meaning understandability was reached.

\textbf{Sensibleness} measured the response appropriateness to what was previously stated by the other dialogue shareholder. A score of 0 means the response did not \textit{follow} what was previously stated or requested, indicating that the current turn-taker ignored the conversation history. Conversely, a score of 1 indicates that the turn-taker acknowledged the conversation history and provided a suitable response.

\textbf{Politeness} measured the courtesy level of each participant towards one another. A score of 0 shows disrespect, discourtesy \textit{inter alia} concerning the remaining participant; score 1 shows the participant was at worst civil and respectful.

The category \textbf{Interaction Quality (IQ)} was adapted from \citet{SCHMITT201512} and scores the turn-taker disposition regarding the previous turn issued by the other dialogue part-taker. This category metric ranges from 1 to 5. With a score of 1, the turn-taker found the previous response to be extremely unsatisfactory; score 2, unsatisfactory; score 3, somewhat unsatisfactory; score 4, somewhat satisfactory; score 5, satisfactory. 

With the above metrics we were able to have a better outlook of the different types of customers and agents, distinguishing behaviour and attitude patterns within a customer support dialogue.

\subsubsection{Dialogue Level Evaluation}

Lastly, we focused on the full dialogue, measuring the conversation in terms of:

\begin{itemize}
  \setlength{\itemsep}{1pt}
  \setlength{\parskip}{0pt}
  \setlength{\parsep}{0pt}
    \item \textbf{Dropped Conversation \{0,1\}}
    \item \textbf{Task Success [1,5]} 
\end{itemize}

\textbf{Dropped Conversation} responds to the questions: \textit{"Was the conversation terminated without a conclusion?"} and/or \textit{"Was the conversation dropped?"}. A score 0 means the conversation reached its end. Conversely, a score of 1 means a dropped conversation, i.e., the conversation did not reach its end, implying that the issue was not resolved.

\textbf{Task Success} dwells with the success of the interaction. This category responds to the following question: \textit{"Was the agent able to fulfil the customer’s request?"} The dialogue success was measured according to the following scores:
\begin{itemize}
  \setlength{\itemsep}{1pt}
  \setlength{\parskip}{0pt}
  \setlength{\parsep}{0pt}
    \item A score of 1 means the agent failed to understand and fulfil the customer’s request;
    \item A score of 2 means the agent understood the request but failed to satisfy it in any way;
    \item A score of 3 means the agent understood the customer’s request and either partially satisfies the request or provided information on how the request can be fulfilled;
    \item A score of 4 means the agent understood and satisfied the customer request, but provided more information than what the customer requested or took unnecessary turns before meeting the request;
    \item A score of 5 means the agent understood and satisfied the customer request completely and efficiently.

\end{itemize}

%With the above described multi-layer annotations, we were able to completely understand and measure the different behavioural and to \textit{shape} the basis for the creation of a future platform with available  automatic translation and adaptive dialog systems that, supported by artificial intelligence technologies, provide assistance to customer support agents in real-time, creating a more successful and seamless customer support experience.

\subsection{Interannotator agreement (IAA)}

Since all annotators were also fluent in European Portuguese (PT-PT), we conducted a trial annotation using 10 dialogues of the corresponding subset to gauge inter-annotator agreement between the annotations. The observed agreement is presented in Table \ref{tab:iaa} \footnote{Due to the class imbalance, regular IAA metrics such as Cohen's kappa \cite{kappacohen} are uninformative.}. Of note, we observe that IQ and Task Success are the annotations that have the lowest agreement, which is expected given the highly subjective nature of these annotations and the fact they are annotated using a Likert Scale. By mapping these annotations to a binary decision (joining the last 2 and 3 ranks together for IQ and Task Success, respectively), the (full/partial) agreement increases to \textbf{(87.4/12.6)} and \textbf{(80.00/20.00)} for IQ and Task Success, respectively.

\section{MAIA Dataset}
\label{sec:description}

\subsection{Statistics}

\begin{table}[h!]
\centering
\small
\begin{tabular}{lc}
\toprule
\textbf{Annotation}           & \textbf{Count}       \\\midrule
\textbf{Correctness \{0,1,2\}}          & 205 | 938 | 23,730         \\
\textbf{Templated \{0,1\}}            & 18,174 | 6,602            \\
\textbf{Engagement \{0,1\}}           & 315 | 23,712             \\
\textbf{Understanding \{0,1\}}        & 136 | 9,470              \\
\textbf{Sensibleness \{0,1\}}         & 127 | 9,478              \\
\textbf{Politeness \{0,1\}}           & 345 | 9,390             \\
\textbf{IQ [1,5]}                   & 89 | 479 | 1,665 | 4,358 | 3,012 \\
\textbf{Dropped Conv. \{0,1\}} & 499 | 112               \\
\textbf{Task Success [1,5]}         & 35 | 63 | 141 | 27 | 347     \\\bottomrule
\end{tabular}
\caption{Statistical information pertaining to the annotations of the MAIA dataset.}
\label{tab:stat_annot}
\end{table}

\begin{figure}[h]
\centering
  \includegraphics[width=0.4\textwidth]{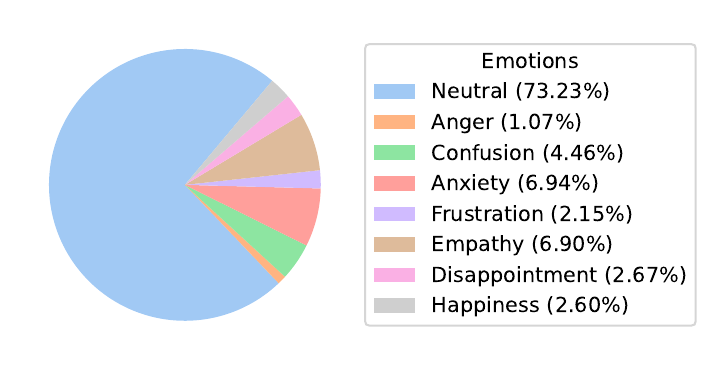}
  \caption{Emotion distribution of the MAIA dataset.}
  \label{fig:emopie}
\end{figure}

The dataset consists of a total of 612 dialogues, split into 5 subsets of different languages and/or companies (identified using a unique integer). Table \ref{tab:stats} presents the statistical information of the dataset and corresponding subsets. Additional statistics on the quality annotations is presented in Table \ref{tab:stat_annot}, with Figure \ref{fig:emopie} illustrating the emotion distribution.

\subsection{Structure}

Whilst the majority of dialogues follows a typical turn-taking approach, we find some instances where one of the participants breaks the flow of the conversation. This occurs when the next turn taker does not respond within an appropriate time frame (according to the other side). This is especially true at the end of the dialogues, where the customer terminates the conversation abruptly, irrespective of whether the issue was resolved. Additionally, these interactions are aided by automated system that responds on behalf of the agent: (1) when the customer doesn't reply within a given time frame, resulting in the system reminding the customer of the ongoing customer support interaction before terminating the conversation; (2) at the end of the dialogues, requesting customer satisfaction survey and providing additional steps, if applicable.

\subsection{Observations and Discussion}

\begin{figure}[h]
\centering
  \includegraphics[width=0.45\textwidth]{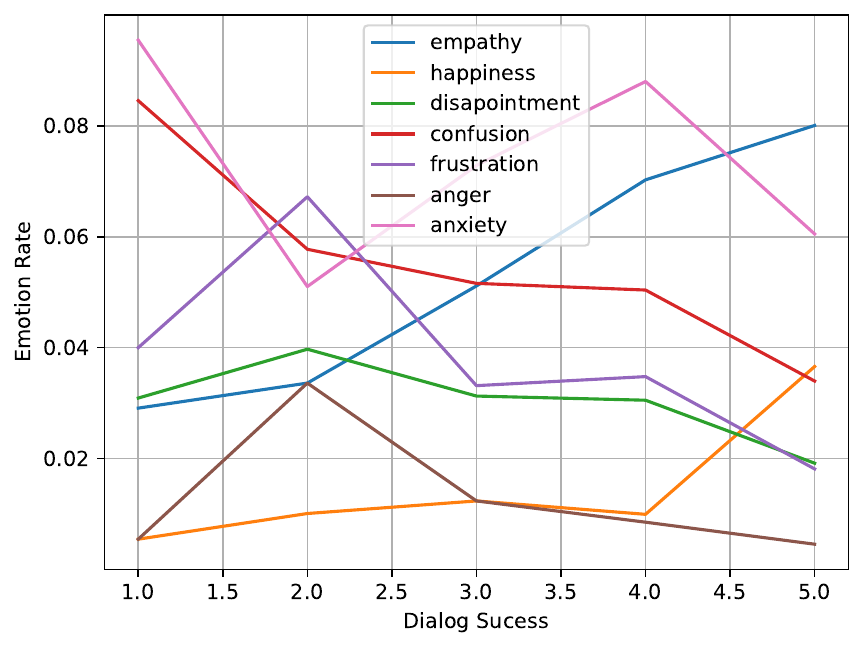}
  \caption{Proportion of non-neutral Emotion Rates across all Dialogue Success levels}
  \label{fig:emotion}
\end{figure}

\paragraph{Emotion correlates with interaction quality and dialogue success.} We hypothesise a positive correlation between emotion and dialogue success levels since the emotions of the interlocutors are related with the outcome of the experiment. This can be observed in Figure \ref{fig:emotion}, where we note a rise in empathy and happiness, together with a decrease in negative emotions.
%being very notorious the rise in empathy across dialogue success levels, a desired emotional display specially for the agent. 
Simultaneously, a positive correlation between emotion and Interaction Quality (IQ) should also be observed. For each turn, we mapped the emotions into a 3 class sentiment (-1,0,1) and report a Pearson and Spearman correlation of \textbf{0.4136} and \textbf{0.5494}, respectively.

\begin{figure}[h]
  \centering
  \includegraphics[width=0.48\textwidth]{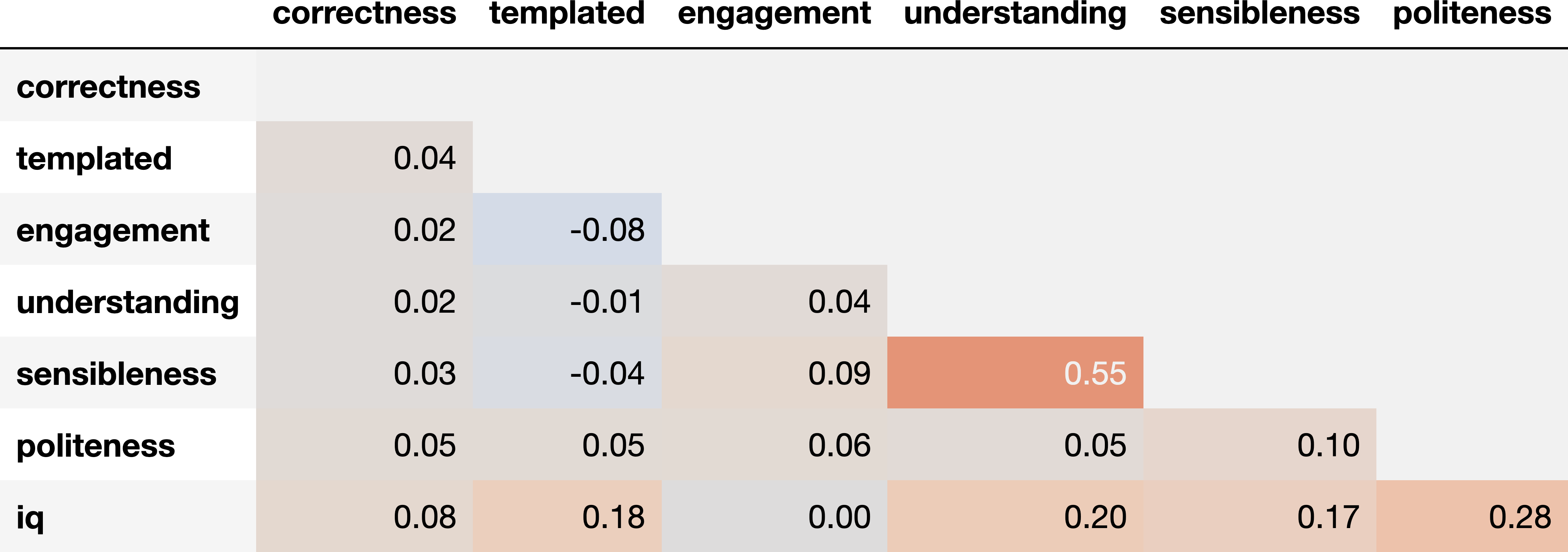}
  \caption{Pairwise Pearson correlation matrix of sentence and turn level annotations.}
  \label{fig:corr}
\end{figure}

\begin{figure*}[t]
  \centering
  \includegraphics[width=.95\textwidth]{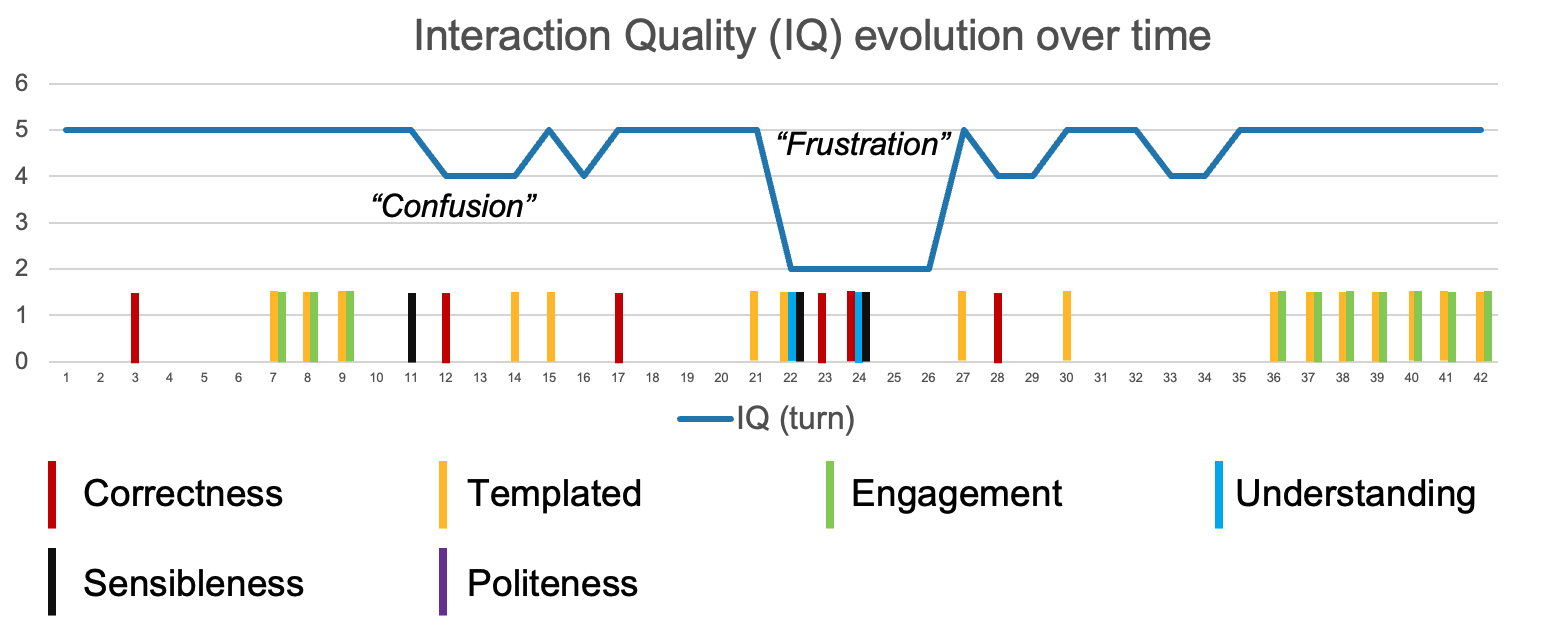}
  \caption{Evolution of the annotation \textit{Interaction Quality} over a dialogue, together with relevant sentence and turn level annotations. Each spike in the lower portion of the figure denotes a negative annotation.}
  \label{fig:iq}
\end{figure*}

\paragraph{Low dialogue subqualities result in loss of customer satisfaction.}

One assumes a decrease in IQ can be attributed, in part, to the occurrence of low quality responses by either participants. Looking at Figure \ref{fig:corr}, subqualities such as \textit{Understanding},  \textit{Sensibleness} and \textit{Politeness} are in fact somewhat correlated with IQ. Engaging responses are uncorrelated with IQ, however. This is likely due to the nature of the dialogue itself, which generally involves the agent dictating steps and/or terms and conditions pertaining to the product, which are verbatim of existing content.

\begin{table*}[t]
 \centering
 \small
  \begin{tabular}{l|cc|cccccccc}
    
     \textbf{Subset} & \textbf{Macro-F1} &
     \textbf{Micro-F1} &
     \textbf{Emp} & 
     \textbf{Hap} & \textbf{Disa} & \textbf{Conf} & \textbf{Frus} & \textbf{Ang} & \textbf{Anx} & \textbf{Neu} \\
    \hline
    All &47.98&83.49&67.16 &45.74 & 34.37&48.59 &22.22&16.96&58.05&90.71 \\
   
    DE-1 &48.26&82.55&72.70 &39.60 &41.13 &42.00 &10.50&31.22&59.31&89.60    \\
    DE-2 &44.59&88.29 &63.28&37.49 & 26.38&53.87 &16.68&8.00&57.98&93.02    \\
    PT\_BR-2 &39.84&83.51&42.61 &53.93 &31.83 &53.64 &25.34&0&20.13&91.27    \\
    PT\_PT-3 &39.27 &81.94&85.33 &21.33 &0&0.5917&31.34&0&0&91.05    \\
    PT-BR-4 &30.50&77.51 &47.14&0 &23.07 &52.78 &26.63&0&6.67&87.74    \\
    
    \hline
  \end{tabular} 
   \caption{Emotion Recognition results for each subset and the full dataset. Results are an average of 5 runs.}
 \label{t7}
\end{table*}

\paragraph{Low quality interactions can be recovered successfully.} 

%The inability of the agent to rectify issues arising from this degradation in the subqualities may also result in lower task success rates, or even dropped conversations. These dynamics can be observed by looking at the subqualities throughout a dialogue. 
%Figure \ref{fig:cm_ts} compares Task Success against Dropped Conversations and IQ of the last turn, where it is clear these annotations can give a good indication of the success of the task. Additionally,
Figure \ref{fig:iq} presents a use case where a decrease of IQ is observed and rectified by the agent, resulting in a positive outcome: Around turn 21 we observe a large degradation in IQ which is paired with frustration. This is a result of the responses by the agent being templated and ineffective to solve the issue at hand. This is further exacerbated due to the lack of understanding between the participants, which is eventually resolved, increasing the quality of the interaction.

\section{Benchmark Evaluation}
\label{sec:benchmark}

Given the focus of the annotation work was on emotions and dialogue quality, in this section we evaluate existing mainstream approaches for emotion recognition and automatic dialogue evaluation.

\subsection{Emotion Recognition in Conversations}

State-of-the-art approaches for Emotion Recognition in Conversations (ERC) produce representations of each sentence using pretrained language models and then model the interactions between these representations with classification modules. Approaches such as leveraging conversational context or speaker specific modelling typically resort to architectures such as gated and graph neural networks \cite{poria2019emotion}.

\subsubsection{Experiments}

For our benchmark, we finetuned a pretrained Encoder model, more specifically XLM-RoBERTa \cite{conneau-etal-2020-unsupervised}. %as it supports the languages included on the MAIA dataset.
We conducted train/dev/test splits at the dialogue level for each subset, employing a distribution of 70\%/10\%/20\%, 
respectively, and ensuring the original distribution of emotion classes on all splits whenever possible.
During training and evaluation, we used the source text 
while considering only the primary emotion labels, disregarding secondary emotion annotations.
Performance is evaluated using Macro, Micro and individual emotion label F1 scores across all languages and the whole dataset.
Additional training details are available in Appendix \ref{sec:app_exp}.

\begin{table*}[t]
\centering
\small
\begin{tabular}{c|l|ccccccc}
                     & \textbf{Model} & \textbf{Correctness} & \textbf{Templated} & \textbf{Engagement} & \textbf{Understanding} & \textbf{Sensibleness} & \textbf{Politeness} & \textbf{IQ}     \\ \hline
\multirow{3}{*}{CST} & VSP            & \textbf{0.6361}      & \textbf{0.6541}    & 0.4667              & 0.5112                 & 0.4943                & 0.5091              & \textbf{0.5307} \\
                     & NSP            & 0.5444               & 0.4645             & \textbf{0.5083}     & \textbf{0.5734}        & \textbf{0.5831}       & \textbf{0.5603}     & 0.4842          \\
                     & ENG            & 0.5205               & 0.5795             & 0.4545              & 0.5374                 & 0.5484                & 0.5510              & 0.4740          \\ \hline
\multirow{3}{*}{AGT} & VSP            & \textbf{0.7061}      & \textbf{0.6073}    & 0.4601              & 0.4648                 & 0.4973                & 0.5165              & \textbf{0.5083} \\
                     & NSP            & 0.5850               & 0.4888             & \textbf{0.5182}     & \textbf{0.5657}        & \textbf{0.5864}       & \textbf{0.5821}     & 0.5029          \\
                     & ENG            & 0.5443               & 0.5794             & 0.4503              & 0.5514                 & 0.5548                & 0.5756              & 0.4742          \\ \hline 
\end{tabular}
\caption{Balanced Accuracy Score of the binary subquality prediction for the MAIA dataset, from the point of view of the CST (customer-LANG) and AGT (agent-EN). Best results for each of them per subquality in \textbf{bold}.}
\label{tab:subquality_Res}
\end{table*}
\subsubsection{Results}

Results for this benchmark are presented in Table
\ref{t7}.
We report a Macro-f1 score of 47.98 for the whole MAIA dataset. This result is within the performance of typical ERC models for other datasets that also have an imbalanced class distribution. The most represented \textit{Neutral} class has a high F1 score across all subsets, heavily influencing the Micro-F1 score. Other well represented classes such as \textit{Empathy} and \textit{Anxiety} also have high F-scores, whereas minority classes have lower scores. In some subsets, individual emotion labels present very low to null F1 scores, again, a result of the class imbalance issues. In fact, due to the limited number of examples for these emotions in some subsets, a handful of missclassifications yield single digit F1 scores.

\subsection{Automatic Dialogue Evaluation}

Most competitive metrics for turn-level dialogue evaluation leverage pretrained Encoder models that are finetuned using well-defined self-supervised tasks \citep{yeh-etal-2021-comprehensive, Zhang2021AutomaticEA}. These approaches generate synthetically negative samples from the original dialogue data, thereby circumventing limitations w.r.t the lack of quality annotated dialogues. However, it isn't clear these approaches extend to task-oriented dialogues and/or Multilingual models, since dialogue data is exclusively open-domain and in English. As such, the MAIA dataset can be used as a benchmark to study these characteristics.

\subsubsection{Experiments}

Similar to approaches mentioned above, we finetuned XLM-RoBERTa for \textbf{ENG} (Engagement) using the ENDEX data \cite{xu-etal-2022-endex}; and \textbf{VSP} (Valid Sentence Prediction) and \textbf{NSP} (Next Sentence Prediction) using self-supervised data generated from DailyDialog \citep{li-etal-2017-dailydialog}. \textbf{VSP} is mostly concerned with the syntactic fluency of the response, which maps to \textit{Correctness} and \textit{Templated}; \textbf{NSP} evaluates textual entailment, which maps to \textit{Understanding} and \textit{Sensibleness}; Finally, since we have Engagement annotations, the evaluation of the ENG submetric is straightforward. The mapping between these submetrics and the remaining annotations is less obvious, but most evaluation frameworks that leverage these submetrics have shown positive correlations with quality aspects that do not map to the submetrics \cite{yeh-etal-2021-comprehensive}.

For this task, we mapped existing sentence-level annotations to turn-level by selecting the minimum of the given turn. For simplicity, we report the Balanced Accuracy Score (BAS), which in this case corresponds to the average recall obtained on the positive (1) and negative (0) classes. The BAS for outputting a single class is 0.5. As such, we consider always outputting the majority class as the baseline. For Correctness, we considered a turn to be positive when all sentences have a score higher than 0; for IQ, only turns with a score of 4 or 5 are labelled positive. We indicate results for both languages, i.e, the context-response pairs from the point of view of the Customer (CST) (original language, with agent text translated) and the Agent (AGT) (in English, customer text translated). Note that, in this case, we conducted zero-shot inference on customer languages using models finetuned only on English data. Additional details available in \ref{sec:app_exp}.

\subsubsection{Results}

For ease of reading, we aggregate the results of all subsets and report the BAS in Table \ref{tab:subquality_Res}. It is clear some models are best suited to predict only some subqualities. However, despite \textbf{ENG} being trained on engagement data, it underperforms NSP on the \textit{Engagement} annotation. This may be related to the training data itself: \textit{Engagement} in the context of open-domain dialogue is different than in customer support. Further, we observe that most models only slightly outperform just predicting the positive class. This means typical approaches for automatic subquality prediction are insufficient to adequately predict low quality responses on the MAIA dataset. 

Comparing the results for AGT against CST we note that the trained models do not consistently outperform on a given language. This may indicate finetuning a multilingual encoder with English dialogue data only achieves reasonable results in a multilingual setting. However, it is important to point out (1) that the agent converses in English; (2) the result that is most sensible to linguistic differences is \textbf{VSP} for \textit{Correctness} (since it looks at the syntax), and here we see that the model underperforms for the other languages.

\section{Conclusions}
\label{sec:conclusions}

This paper presents a comprehensive emotion and dialogue quality annotation for the MAIA dataset, a collection of genuine bilingual customer support conversations. All in all, we annotate 612 dialogues amounting to over 24k sentences. Besides allowing for an opportunity to study the dynamics of Machine Translation aided customer support conversations, it also provides a novel opportunity to benchmark and explore applications of existing and future NLP models applied to dialogue.

Results on the different benchmarks indicate there is still room for improving existing models. LLMs such as GPT-4 \cite{openai2023gpt4} show impressive classification and generation capabilities, and may prove useful in augmenting existing customer support datasets to new languages and tasks. These in turn can be used to build data-driven classifiers or end-to-end conversational agents that are robust to new languages and domains.

\section*{Limitations}
\label{sec:limitations}

Perhaps the main limitation of this work concerns the lack of several annotators for each subset. Even with well defined guidelines, individual biases may affect the annotations, especially for dialogue quality as it is highly subjective \cite{smith-etal-2022-human}. By having several annotators evaluate the conversations, one could've leveraged "the wisdom of the crowd", but this approach also comes with its own limitations \cite{jain2010investigation}. Ideally we would've employed several expert annotators, but were only able to recruit a single expert for each language. In any case, we conducted a trial annotation where all annotators participated and report moderate to strong agreement on a subset of the dataset.

Another limitation pertains to the dataset itself. Despite being structured and evaluated as a dyadic interaction, the actual conversations may not follow this structure. For instance, whenever one of the participants takes to long to respond, the other may follow-up on its original turn with a reminder. Given we do not have access to this temporal information, these sentences were lumped together into a single turn. Also pertaining to metadata information is the lack of the original customer support guidelines. This makes the \textit{Templated} annotation a subjective observation from the point of view of the customer. However, since we are framing this annotation from a quality perspective, we believe our annotation accurately reflects the perception of quality from the P.O.V of the customer.

\section*{Ethics Statement}
\label{sec:ethics}

This work leverages real world dialogues. A comprehensive anonymization process was conducted to ensure all PII were removed, in accordance with EU's GDPR. The annotations were conducted exclusively by highly-educated European Portuguese, which were paid a fair wage according to local costs of living. Despite being native speakers of the languages they evaluated, one might argue notions of quality are strongly tied to the culture and not the language. As such, they may not accurately represent other groups.

\section*{Acknowledgements}

This research was supported by the Portuguese Recovery and Resilience Plan through project C645008882-00000055 (Responsible.AI), by national funds through \textit{Fundação para a Ciência e a Tecnologia} (FCT) with references PRT/BD/152198/2021, UI/BD/154561/2022, 2022.12091.BD, and UIDB/50021/2020, by the P2020 program MAIA (LISBOA-01-0247-FEDER-045909), and by the EU’s Horizon Europe (UTTER, HORIZON- CL4-2021-HUMAN-01-13, contract 101070631). We also thank the reviewers for their constructive feedback.

\bibliography{anthology,custom}
\bibliographystyle{acl_natbib}

\appendix

\section{Experimental Setup}
\label{sec:app_exp}

All experiments used XLM-RoBERTa-large downloaded from the Transformers library by Hugging Face \footnote{\url{huggingface.co/xlm-roberta-large}}. All parameters were trained/finetuned using Adam optimizer \cite{DBLP:journals/corr/KingmaB14} and a single Quadro RTX 6000 24GB GPU for all experiments was used.

\subsection{Emotion Recognition in Conversations} 

\paragraph{Training and Hyperparameters}
We trained XLM-R with the cross-entropy loss with logits. An initial learning rate of 1e-5 and 5e-5 was used for the encoder and the classification head, respectively, with a layer-wise decay rate
of 0.95 after each training epoch for the encoder, which was frozen for the first epoch. The batch size was set to 4 and Gradient clipping to 1.0. Early stopping was used to terminate training if there was no improvement after 5 consecutive epochs on the validation set over macro-F1, for a maximum of 10 epochs. The best performing model on the validation set was selected for testing.

\subsection{Dialogue Evaluation} 

\paragraph{Processing} For the dialogue data preprocessing we used spaCy \footnote{\url{spacy.io}}. In this paper, we followed the approach used by \citet{phy-etal-2020-deconstruct} and initially proposed by \citet{sinha-etal-2020-learning}. In detail, we train models trained to differentiate between positive samples and synthetic negative samples from DailyDialog \cite{li-etal-2017-dailydialog}: For the \textbf{VSP} model, \textbf{Positive} samples are perturbed by randomly applying one of the following: (1) no perturbation, (2) punctuation removal, (3) stop-word removal. \textbf{Negative} samples are generated by randomly applying one of the following rules: (1) word reorder (shuffling the ordering of the words); (2) word-drop; and (3) word-repeat (randomly repeating words). For the \textbf{NSP} model, \textbf{positive} responses are drawn directly from the dialog; \textbf{negative} responses are randomly selected and a token coverage test discards semantically similar sentences. All responses are processed using the positive-sample heuristic used by VSP. The \textbf{ENG} model was trained directly on the 80k split with negative sampled data of the ENDEX dataset \cite{xu-etal-2022-endex}.

\paragraph{Training and Hyperparameters} All models were obtained following the recipe from \citet{mendoncaetal2023towards}. In detail, a token representing the speaker was added for each turn, and a history length of 3 turns was used. We applied a regression head consisting of a 2-layer MLP with a hidden size of 1024 and a hyperbolic tangent function as activation for prediction. A learning rate of 3e-6 for 3 epochs using a batch size of 16 was used. Evaluation was conducted every 10,000 steps. The best performing model on the evaluation set was selected for testing.

\section{Example Dialogue}
\label{sec:dial}

% \usepackage{multirow}
\begin{table*}[t]
\scriptsize
\centering
\rotatebox{90}{

\begin{tabular}{ll|cccc|cccc|cc}
\toprule \\
\multicolumn{2}{c|}{\textbf{Text}}                                                                                                                                                                                                                                                                                                                                                                                                          & \multicolumn{4}{c|}{\textbf{Sentence}}                                                                                                        & \multicolumn{4}{c|}{\textbf{Turn}}                                                                                                                                                    & \multicolumn{2}{c}{\textbf{Dialogue}}       \\ \hline
\multicolumn{1}{l|}{\textbf{SRC}}                                                                                                                                                                                                & \textbf{MT}                                                                                                                                                                                              & \multicolumn{1}{c|}{\textbf{Emo}}                        & \textbf{Cor}              & \textbf{Tem}              & \textbf{Eng}               & \textbf{Und}                                & \textbf{Sen}                                & \textbf{IQ}                                 & \textbf{Pol}                                & \textbf{DC}          & \textbf{TS}          \\ \hline
\multicolumn{1}{l|}{\cellcolor[HTML]{ECF4FF}\begin{tabular}[c]{@{}l@{}}Oii, fui cobrada por um plano\\ q n estou usando, e estou \\ solicitando retorno\end{tabular}}                                                            & \cellcolor[HTML]{ECF4FF}\begin{tabular}[c]{@{}l@{}}Oii, I was charged for a plan q n am using, and I am\\ requesting return\end{tabular}                                                                 & \multicolumn{1}{c|}{\cellcolor[HTML]{ECF4FF}Frustratiom} & \cellcolor[HTML]{ECF4FF}0 & \cellcolor[HTML]{ECF4FF}0 & \cellcolor[HTML]{ECF4FF}NA & \cellcolor[HTML]{ECF4FF}NA                  & \cellcolor[HTML]{ECF4FF}NA                  & \cellcolor[HTML]{ECF4FF}NA                  & \cellcolor[HTML]{ECF4FF}1                   &                      &                      \\ \cline{1-10}
\multicolumn{1}{l|}{\cellcolor[HTML]{FDEDEC}Good day :)}                                                                                                                                                                         & \cellcolor[HTML]{FDEDEC}Tenha um bom dia!                                                                                                                                                                & \multicolumn{1}{c|}{\cellcolor[HTML]{FDEDEC}Neutral}     & \cellcolor[HTML]{FDEDEC}2 & \cellcolor[HTML]{FDEDEC}1 & \cellcolor[HTML]{FDEDEC}1  & \cellcolor[HTML]{FDEDEC}                    & \cellcolor[HTML]{FDEDEC}                    & \cellcolor[HTML]{FDEDEC}                    & \cellcolor[HTML]{FDEDEC}                    &                      &                      \\ \cdashline{1-6}
\multicolumn{1}{l|}{\cellcolor[HTML]{FDEDEC}\begin{tabular}[c]{@{}l@{}}Thanks for reaching out and\\ I'm sorry for any confusion here!\end{tabular}}                                                                             & \cellcolor[HTML]{FDEDEC}\begin{tabular}[c]{@{}l@{}}Obrigado por entrar em contato e desculpe-me por\\ qualquer confusão aqui!\end{tabular}                                                               & \multicolumn{1}{c|}{\cellcolor[HTML]{FDEDEC}Neutral}     & \cellcolor[HTML]{FDEDEC}2 & \cellcolor[HTML]{FDEDEC}1 & \cellcolor[HTML]{FDEDEC}1  & \cellcolor[HTML]{FDEDEC}                    & \cellcolor[HTML]{FDEDEC}                    & \cellcolor[HTML]{FDEDEC}                    & \cellcolor[HTML]{FDEDEC}                    &                      &                      \\\cdashline{1-6}
\multicolumn{1}{l|}{\cellcolor[HTML]{FDEDEC}\begin{tabular}[c]{@{}l@{}}\#PRS\_ORG\# memberships\\ renew each month unless you\\ request a cancellation via one\\ of our agents.\end{tabular}}                                    & \cellcolor[HTML]{FDEDEC}\begin{tabular}[c]{@{}l@{}}As assinaturas \#PRS\_ORG\# são renovadas todos os meses,\\ a menos que você solicite um cancelamento através de um\\ de nossos agentes.\end{tabular} & \multicolumn{1}{c|}{\cellcolor[HTML]{FDEDEC}Neutral}     & \cellcolor[HTML]{FDEDEC}2 & \cellcolor[HTML]{FDEDEC}1 & \cellcolor[HTML]{FDEDEC}1  & \cellcolor[HTML]{FDEDEC}                    & \cellcolor[HTML]{FDEDEC}                    & \cellcolor[HTML]{FDEDEC}                    & \cellcolor[HTML]{FDEDEC}                    &                      &                      \\\cdashline{1-6}
\multicolumn{1}{l|}{\cellcolor[HTML]{FDEDEC}\begin{tabular}[c]{@{}l@{}}It seems that our system did not\\ process a request from you before\\ your cycle renewed which is why\\ you were charged.\end{tabular}}                  & \cellcolor[HTML]{FDEDEC}\begin{tabular}[c]{@{}l@{}}Parece que nosso sistema não processou uma solicitação\\ de você antes da renovação do seu ciclo, e é por isso\\ que você foi cobrado.\end{tabular}   & \multicolumn{1}{c|}{\cellcolor[HTML]{FDEDEC}Neutral}     & \cellcolor[HTML]{FDEDEC}2 & \cellcolor[HTML]{FDEDEC}0 & \cellcolor[HTML]{FDEDEC}1  & \cellcolor[HTML]{FDEDEC}                    & \cellcolor[HTML]{FDEDEC}                    & \cellcolor[HTML]{FDEDEC}                    & \cellcolor[HTML]{FDEDEC}                    &                      &                      \\\cdashline{1-6}
\multicolumn{1}{l|}{\cellcolor[HTML]{FDEDEC}\begin{tabular}[c]{@{}l@{}}Unfortunately we are not able to\\ refund \#PRS\_ORG\# memberships\\ retroactively.\end{tabular}}                                                         & \cellcolor[HTML]{FDEDEC}\begin{tabular}[c]{@{}l@{}}Infelizmente, não podemos reembolsar as assinaturas\\ da \#PRS\_ORG\# retroativamente.\end{tabular}                                                   & \multicolumn{1}{c|}{\cellcolor[HTML]{FDEDEC}Neutral}     & \cellcolor[HTML]{FDEDEC}2 & \cellcolor[HTML]{FDEDEC}0 & \cellcolor[HTML]{FDEDEC}1  & \cellcolor[HTML]{FDEDEC}                    & \cellcolor[HTML]{FDEDEC}                    & \cellcolor[HTML]{FDEDEC}                    & \cellcolor[HTML]{FDEDEC}                    &                      &                      \\\cdashline{1-6}
\multicolumn{1}{l|}{\cellcolor[HTML]{FDEDEC}\begin{tabular}[c]{@{}l@{}}Our refund policy is stated in our\\ Help Center here for further reference\\ \#URL\#\end{tabular}}                                                       & \cellcolor[HTML]{FDEDEC}\begin{tabular}[c]{@{}l@{}}Nossa política de reembolso está descrita em nosso\\ centro de ajuda aqui para mais referências: \#URL\#\end{tabular}                                 & \multicolumn{1}{c|}{\cellcolor[HTML]{FDEDEC}Neutral}     & \cellcolor[HTML]{FDEDEC}2 & \cellcolor[HTML]{FDEDEC}1 & \cellcolor[HTML]{FDEDEC}1  & \multirow{-6}{*}{\cellcolor[HTML]{FDEDEC}1} & \multirow{-6}{*}{\cellcolor[HTML]{FDEDEC}1} & \multirow{-6}{*}{\cellcolor[HTML]{FDEDEC}5} & \multirow{-6}{*}{\cellcolor[HTML]{FDEDEC}1} &                      &                      \\ \cline{1-10}
\multicolumn{1}{l|}{\cellcolor[HTML]{ECF4FF}\begin{tabular}[c]{@{}l@{}}E tem como eu pssar a assinatura pra\\ outra pessoa???\end{tabular}}                                                                                      & \cellcolor[HTML]{ECF4FF}And how do I get the signature for someone else???                                                                                                                               & \multicolumn{1}{c|}{\cellcolor[HTML]{ECF4FF}Frustration} & \cellcolor[HTML]{ECF4FF}1 & \cellcolor[HTML]{ECF4FF}0 & \cellcolor[HTML]{ECF4FF}1  & \cellcolor[HTML]{ECF4FF}                    & \cellcolor[HTML]{ECF4FF}                    & \cellcolor[HTML]{ECF4FF}                    & \cellcolor[HTML]{ECF4FF}                    &                      &                      \\\cdashline{1-6}
\multicolumn{1}{l|}{\cellcolor[HTML]{ECF4FF}\begin{tabular}[c]{@{}l@{}}Pq meu dinheiro n é de graça pra ser\\ gasto a toa\end{tabular}}                                                                                          & \cellcolor[HTML]{ECF4FF}Pq my money n is for free to be spent for nothing                                                                                                                                & \multicolumn{1}{c|}{\cellcolor[HTML]{ECF4FF}Frustration} & \cellcolor[HTML]{ECF4FF}0 & \cellcolor[HTML]{ECF4FF}0 & \cellcolor[HTML]{ECF4FF}1  & \multirow{-2}{*}{\cellcolor[HTML]{ECF4FF}1} & \multirow{-2}{*}{\cellcolor[HTML]{ECF4FF}1} & \multirow{-2}{*}{\cellcolor[HTML]{ECF4FF}2} & \multirow{-2}{*}{\cellcolor[HTML]{ECF4FF}0} &                      &                      \\ \cline{1-10}
\multicolumn{1}{l|}{\cellcolor[HTML]{FDEDEC}We apologize for the frustration here.}                                                                                                                                              & \cellcolor[HTML]{FDEDEC}Lamentamos pelo inconveniente.                                                                                                                                                   & \multicolumn{1}{c|}{\cellcolor[HTML]{FDEDEC}Neutral}     & \cellcolor[HTML]{FDEDEC}2 & \cellcolor[HTML]{FDEDEC}1 & \cellcolor[HTML]{FDEDEC}1  & \cellcolor[HTML]{FDEDEC}                    & \cellcolor[HTML]{FDEDEC}                    & \cellcolor[HTML]{FDEDEC}                    & \cellcolor[HTML]{FDEDEC}                    &                      &                      \\\cdashline{1-6}
\multicolumn{1}{l|}{\cellcolor[HTML]{FDEDEC}\begin{tabular}[c]{@{}l@{}}As an exception given the circumstances,\\ I refunded your most recent membership\\ charged, and I cancelled your membership\\ immediately.\end{tabular}} & \cellcolor[HTML]{FDEDEC}\begin{tabular}[c]{@{}l@{}}Como exceção, dadas as circunstâncias, reembolsei\\ a sua assinatura mais recente cobrada e cancelei sua\\ assinatura imediatamente.\end{tabular}     & \multicolumn{1}{c|}{\cellcolor[HTML]{FDEDEC}Neutral}     & \cellcolor[HTML]{FDEDEC}2 & \cellcolor[HTML]{FDEDEC}0 & \cellcolor[HTML]{FDEDEC}1  & \cellcolor[HTML]{FDEDEC}                    & \cellcolor[HTML]{FDEDEC}                    & \cellcolor[HTML]{FDEDEC}                    & \cellcolor[HTML]{FDEDEC}                    &                      &                      \\\cdashline{1-6}
\multicolumn{1}{l|}{\cellcolor[HTML]{FDEDEC}\begin{tabular}[c]{@{}l@{}}You can expect this refund to arrive in 5-7\\ days depending on your bank/carrier, and\\ you won't be charged again moving forward.\end{tabular}}         & \cellcolor[HTML]{FDEDEC}\begin{tabular}[c]{@{}l@{}}O reembolso estará disponível em 5 a 7 dias,\\ dependendo do seu banco, e não haverá mais\\ cobranças.\end{tabular}                                   & \multicolumn{1}{c|}{\cellcolor[HTML]{FDEDEC}Neutral}     & \cellcolor[HTML]{FDEDEC}2 & \cellcolor[HTML]{FDEDEC}0 & \cellcolor[HTML]{FDEDEC}1  & \cellcolor[HTML]{FDEDEC}                    & \cellcolor[HTML]{FDEDEC}                    & \cellcolor[HTML]{FDEDEC}                    & \cellcolor[HTML]{FDEDEC}                    &                      &                      \\\cdashline{1-6}
\multicolumn{1}{l|}{\cellcolor[HTML]{FDEDEC}\begin{tabular}[c]{@{}l@{}}In the meantime, you can view the refunded\\ charge on the billing page in your Account Settings.\end{tabular}}                                           & \cellcolor[HTML]{FDEDEC}\begin{tabular}[c]{@{}l@{}}Enquanto isso, você pode visualizar a cobrança\\ reembolsada na página de faturamento nas suas\\ Configurações da conta.\end{tabular}                 & \multicolumn{1}{c|}{\cellcolor[HTML]{FDEDEC}Neutral}     & \cellcolor[HTML]{FDEDEC}2 & \cellcolor[HTML]{FDEDEC}0 & \cellcolor[HTML]{FDEDEC}1  & \cellcolor[HTML]{FDEDEC}                    & \cellcolor[HTML]{FDEDEC}                    & \cellcolor[HTML]{FDEDEC}                    & \cellcolor[HTML]{FDEDEC}                    &                      &                      \\\cdashline{1-6}
\multicolumn{1}{l|}{\cellcolor[HTML]{FDEDEC}\begin{tabular}[c]{@{}l@{}}Please let me know if you have any other\\ questions or if there is anything else that\\ I can help with.\end{tabular}}                                   & \cellcolor[HTML]{FDEDEC}\begin{tabular}[c]{@{}l@{}}Se tiver outra dúvida ou se precisar de ajuda, é\\ só avisar!\end{tabular}                                                                            & \multicolumn{1}{c|}{\cellcolor[HTML]{FDEDEC}Neutral}     & \cellcolor[HTML]{FDEDEC}2 & \cellcolor[HTML]{FDEDEC}1 & \cellcolor[HTML]{FDEDEC}1  & \multirow{-5}{*}{\cellcolor[HTML]{FDEDEC}1} & \multirow{-5}{*}{\cellcolor[HTML]{FDEDEC}1} & \multirow{-5}{*}{\cellcolor[HTML]{FDEDEC}5} & \multirow{-5}{*}{\cellcolor[HTML]{FDEDEC}1} &                      &                      \\ \cline{1-10}
\multicolumn{1}{l|}{\cellcolor[HTML]{ECF4FF}Muito obrigada!!}                                                                                                                                                                    & \cellcolor[HTML]{ECF4FF}Thank you very much!                                                                                                                                                             & \multicolumn{1}{c|}{\cellcolor[HTML]{ECF4FF}Neutral}     & \cellcolor[HTML]{ECF4FF}2 & \cellcolor[HTML]{ECF4FF}0 & \cellcolor[HTML]{ECF4FF}1  & \cellcolor[HTML]{ECF4FF}1                   & \cellcolor[HTML]{ECF4FF}1                   & \cellcolor[HTML]{ECF4FF}5                   & \cellcolor[HTML]{ECF4FF}1                   &                      &                      \\ \cline{1-10}
\multicolumn{1}{l|}{\cellcolor[HTML]{FDEDEC}My pleasure.}                                                                                                                                                                        & \cellcolor[HTML]{FDEDEC}O prazer é meu.                                                                                                                                                                  & \multicolumn{1}{c|}{\cellcolor[HTML]{FDEDEC}Neutral}     & \cellcolor[HTML]{FDEDEC}2 & \cellcolor[HTML]{FDEDEC}0 & \cellcolor[HTML]{FDEDEC}1  & \cellcolor[HTML]{FDEDEC}                    & \cellcolor[HTML]{FDEDEC}                    & \cellcolor[HTML]{FDEDEC}                    & \cellcolor[HTML]{FDEDEC}                    &                      &                      \\\cdashline{1-6}
\multicolumn{1}{l|}{\cellcolor[HTML]{FDEDEC}Is there anything else I can help with?}                                                                                                                                             & \cellcolor[HTML]{FDEDEC}Posso ajudar com mais alguma coisa?                                                                                                                                              & \multicolumn{1}{c|}{\cellcolor[HTML]{FDEDEC}Neutral}     & \cellcolor[HTML]{FDEDEC}2 & \cellcolor[HTML]{FDEDEC}1 & \cellcolor[HTML]{FDEDEC}1  & \cellcolor[HTML]{FDEDEC}                    & \cellcolor[HTML]{FDEDEC}                    & \cellcolor[HTML]{FDEDEC}                    & \cellcolor[HTML]{FDEDEC}                    &                      &                      \\\cdashline{1-6}
\multicolumn{1}{l|}{\cellcolor[HTML]{FDEDEC}\begin{tabular}[c]{@{}l@{}}It seems like you're busy right now, so\\ I'm going to close out the chat.\end{tabular}}                                                                  & \cellcolor[HTML]{FDEDEC}\begin{tabular}[c]{@{}l@{}}Parece que você está ocupado agora, então eu\\ vou fechar o chat.\end{tabular}                                                                        & \multicolumn{1}{c|}{\cellcolor[HTML]{FDEDEC}Neutral}     & \cellcolor[HTML]{FDEDEC}2 & \cellcolor[HTML]{FDEDEC}1 & \cellcolor[HTML]{FDEDEC}1  & \cellcolor[HTML]{FDEDEC}                    & \cellcolor[HTML]{FDEDEC}                    & \cellcolor[HTML]{FDEDEC}                    & \cellcolor[HTML]{FDEDEC}                    &                      &                      \\\cdashline{1-6}
\multicolumn{1}{l|}{\cellcolor[HTML]{FDEDEC}\begin{tabular}[c]{@{}l@{}}If you have any other questions or want\\ to get back in contact with us, you can\\ do so here: \#URL\#\end{tabular}}                                     & \cellcolor[HTML]{FDEDEC}\begin{tabular}[c]{@{}l@{}}Se você tiver outras perguntas ou quiser entrar\\ em contato conosco, pode fazê-lo aqui: \#URL\#\end{tabular}                                         & \multicolumn{1}{c|}{\cellcolor[HTML]{FDEDEC}Neutral}     & \cellcolor[HTML]{FDEDEC}2 & \cellcolor[HTML]{FDEDEC}1 & \cellcolor[HTML]{FDEDEC}1  & \cellcolor[HTML]{FDEDEC}                    & \cellcolor[HTML]{FDEDEC}                    & \cellcolor[HTML]{FDEDEC}                    & \cellcolor[HTML]{FDEDEC}                    &                      &                      \\\cdashline{1-6}
\multicolumn{1}{l|}{\cellcolor[HTML]{FDEDEC}Have a great day!}                                                                                                                                                                   & \cellcolor[HTML]{FDEDEC}Tenha um ótimo dia :)                                                                                                                                                            & \multicolumn{1}{c|}{\cellcolor[HTML]{FDEDEC}Neutral}     & \cellcolor[HTML]{FDEDEC}2 & \cellcolor[HTML]{FDEDEC}1 & \cellcolor[HTML]{FDEDEC}1  & \multirow{-5}{*}{\cellcolor[HTML]{FDEDEC}1} & \multirow{-5}{*}{\cellcolor[HTML]{FDEDEC}1} & \multirow{-5}{*}{\cellcolor[HTML]{FDEDEC}5} & \multirow{-5}{*}{\cellcolor[HTML]{FDEDEC}1} & \multirow{-20}{*}{1} & \multirow{-20}{*}{5} \\ \hline
\end{tabular}
}
\caption{Example of a full dialogue extracted from PT\_PT-3. The blue and red shaded rows correspond to turns belonging to the Customer and Agent, respectively.}
\label{tab:ex}
\end{table*}

\end{document}